\documentclass[10pt,twocolumn,letterpaper]{article}

\usepackage{cvpr}              % To produce the CAMERA-READY version

\usepackage{xcolor}         % colors
\usepackage{tabularx, multirow, multicol, graphicx, makecell, enumitem, color, soul, colortbl, adjustbox, amsmath, subcaption, pifont, amssymb}
\usepackage[linesnumbered,ruled,vlined]{algorithm2e}

\definecolor{cvprblue}{rgb}{0.21,0.49,0.74}
\usepackage[pagebackref,breaklinks,colorlinks,allcolors=cvprblue]{hyperref}
\usepackage{fontawesome5}

\usepackage{tikz}

\def\paperID{***} % *** Enter the Paper ID here
\def\confName{CVPR}
\def\confYear{2026}

\title{UVU: Improving Multimodal Understanding via Vision-Language Unified Autoregressive Paradigm}

\begin{document}

\author{%
    \parbox{\textwidth}{ % Use parbox to contain content and allow manual line breaks
        \centering % Center the entire author block
        \textbf{Zhehan Kan}\textsuperscript{1, 2 *} 
        \textbf{Xinghua Jiang}\textsuperscript{2 *}
        \textbf{Yubo Zhu}\textsuperscript{3, 2 *} 
        \textbf{Yanlin Liu}\textsuperscript{1} 
        \textbf{Xiaochen Yang}\textsuperscript{4}
        \textbf{Zhixiang Wei}\textsuperscript{2} \\
        \textbf{Shifeng Liu}\textsuperscript{2} 
        \textbf{Qingmin Liao}\textsuperscript{1} 
        \textbf{Wenming Yang}\textsuperscript{1\textdagger} 
        \textbf{Xin Li}\textsuperscript{2\textdagger} 
        \textbf{Yinsong Liu}\textsuperscript{2} 
        \textbf{Deqiang Jiang}\textsuperscript{2} 
        \textbf{Xing Sun}\textsuperscript{2}
    }
    \\[1.5ex] % Adjust vertical space between authors and affiliations
    \parbox{\textwidth}{ % Another parbox for affiliations
        \centering % Center affiliations
        \textsuperscript{1}Tsinghua University \quad
        \textsuperscript{2}Tencent Youtu Lab \quad
        \textsuperscript{3}Nanjing University \quad
        \textsuperscript{4}University of Glasgow
    }
}

\maketitle

\begin{tikzpicture}[remember picture, overlay]
\node[below right, anchor=south west, xshift=1in, yshift=0.2in, text width=\dimexpr\textwidth-2em\relax] at (current page.south west) {
    % Using a minipage to control content width and ensure left-alignment
    \begin{minipage}[t]{\dimexpr\textwidth-2em\relax}
    \raggedright % Left-align the content within the minipage
    \small % Make the text slightly smaller, common for footnotes
    \textsuperscript{*} Equal contribution. Work done during Zhehan Kan and Yubo Zhu's internship at Tencent Youtu Lab. \\
    \textsuperscript{\textdagger} Corresponding author
    \end{minipage}
};
\end{tikzpicture}

\begin{abstract}
Despite remarkable advancements in multimodal large language models (MLLMs), their fine-grained visual understanding is constrained by a primary reliance on sparse textual supervision. Existing efforts to introduce visual supervision typically do so during post-training, when visual representations have already been largely fixed, causing such signals to act mainly as auxiliary constraints rather than as a primary force for shaping perceptual features. In this paper, we aim to fundamentally reshape the model’s perceptual backbone by incorporating vision supervision directly into the pre-training stage. We observe that pixel-level image patches and textual tokens naturally coexist in a shared, raw high-dimensional space characterized by an inherent input symmetry. Leveraging this insight, we propose \textbf{UVU}, a novel vision-language unified autoregressive framework that eschews vector quantization. It uniquely employs continuous visual encoding for lossless representation of visual inputs and proposes a large-scale iterative hierarchical clustering algorithm to construct a pixel-level visual codebook, thereby extending the vocabulary for unified supervision and enabling autoregressive generation of pixel-level image tokens alongside textual tokens. UVU effectively synergizes pixel-level visual perception with semantic-level visual understanding, internalizing visual reconstruction capabilities and\textbf{ unlocking the facilitative role of visual supervision in enhancing understanding in the pre-training stage}. Extensive experiments across multiple tasks demonstrate that MLLMs are capable of achieving superior multimodal understanding performance under the supervised learning paradigm of UVU.
\end{abstract}    
\section{Introduction}
\label{sec:intro}

\begin{figure}[htbp]
  \centering
  \includegraphics[width=1.0\linewidth]{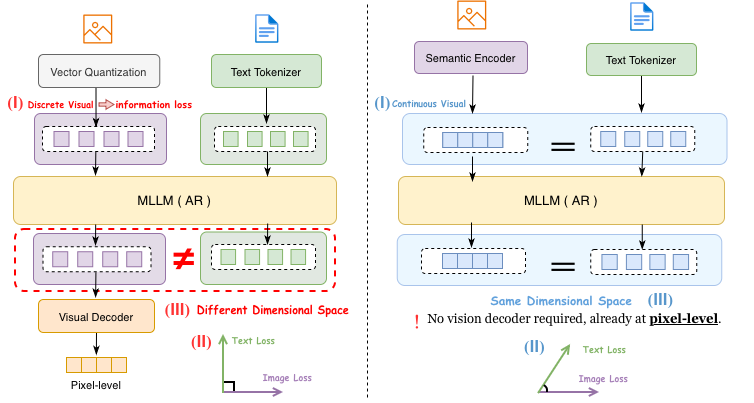}
  \caption{Comparison between traditional AR+VQ architectures (\textbf{left}), which suffer from visual input discretization, image-text loss gradient orthogonality, and external decoder reliance, and the proposed UVU framework (\textbf{right}), which employs continuous visual encoding and pixel-level visual codebooks to unify high-dimensional spaces, internalize visual reconstruction, and enhance multimodal understanding.}
  \label{intro}
\end{figure}

MLLMs have exhibited remarkable advancements in recent years, demonstrating strong, robust, and scalable performance across a diverse range of multimodal tasks and applications~\cite{bai2025qwen25vltechnicalreport, bai2023qwenvlversatilevisionlanguagemodel, chen2024internvlscalingvisionfoundation, liu2023visualinstructiontuning, kan2025tacothinkanswerconsistencyoptimized, zhu2025llmknowsestimatingllmperceived}. Despite their impressive capabilities in general scenarios, MLLMs still encounter limitations in tasks requiring rich visual perception. In most existing approaches, visual representations are primarily learned under text-centric supervision, making it difficult to provide direct and fine-grained visual guidance during training. Therefore, explicit visual supervision offers a more direct mechanism to inject such fine-grained perceptual signals.

To this end, several studies have attempted to enhance the visual understanding capability of MLLMs by introducing vision supervision into vision-language training. In the context of MLLM training, ~\cite{li2025vistaenhancingvisiontextalignment} incorporates visual supervision through vision-text alignment, while ~\cite{wang2024reconstructivevisualinstructiontuning, yoon2025visualrepresentationalignmentmultimodal, li2025spatialforcingimplicitspatial, li2025unleashingintrinsicvisualrepresentation, wang2026autoregressivesemanticvisualreconstruction} leverage visual supervision by predicting or reconstructing image features produced by a vision tokenizer. However, these methods mainly introduce visual supervision at the post-training stage, where the model's visual representations have already been largely shaped by earlier large-scale pre-training. As a result, the visual signal typically serves only as an auxiliary constraint for alignment or reconstruction, rather than a primary force for reshaping fine-grained perceptual features, which limits its effectiveness. Moreover, whether vision supervision can be effectively incorporated into the pre-training stage to fundamentally shape visual representations remains largely unexplored.

In this paper, we aim to fundamentally shape the model’s visual representations from the outset, thereby enhancing its core perceptual and understanding capabilities, by incorporating explicit visual supervision into the pretraining stage. A natural way to realize such supervision is to introduce structured, fine-grained pixel-level visual targets that explicitly encode local perceptual information. In this regard, vector-quantized (VQ) visual tokens offer a robust and prevalent form of discrete supervision by transforming continuous visual features into learnable symbolic units. By supervising these VQ tokens at the output stage of MLLMs, the model is inherently driven to preserve and reason over fine-grained visual details. We begin by investigating the limitations of directly employing pixel-level quantized tokens as a supervisory signal, eg.VQVAE~\cite{oord2018neuraldiscreterepresentationlearning}. As illustrated in the Figure~\ref{intro} left, we identify three key issues: (i) Discretization of input visual features leading to information loss, particularly in high-level semantic details, thereby constraining the richness of continuous representations; (ii) The placement of visual reconstruction's image tokens in post-VQ low-dimensional spaces, contrasted with language generation's textual tokens in high-dimensional semantic spaces, resulting in orthogonality between image loss gradients (emphasizing low-dimensional feature reconstruction) and text loss gradients (prioritizing high-dimensional semantic consistency), which induces conflicting optimization paths and impedes multimodal alignment and integration; (iii) Dependence on external visual decoders for pixel-level visual reconstruction, externalizing visual reconstruction capabilities and fragmenting internal visual understanding in MLLMs, thus preventing end-to-end learning. Nevertheless, we discover that pixel-level image patches and textual tokens coexist in raw high-dimensional spaces with inherent input symmetry. Furthermore, supervising visual reconstruction's image tokens in the pixel-level image patch space can naturally leverage this symmetry to eliminate dependence on VQ encoders at the input and visual decoders at the output. 

Motivated by the aboved insight, we propose a novel vision-language unified autoregressive framework, \textbf{UVU}, as illustrated in the Figure~\ref{intro} right, which eschews VQ altogether. At the input, it uniquely employs continuous visual encoding to preserve lossless representations. We introduce a large-scale iterative hierarchical clustering algorithm that, through iterative deduplication, hierarchical sampling, and clustering training, yields pixel-level image patch cluster centers with high coverage and precision. These centers serve as codewords to assemble a pixel-level visual codebook, extending the vocabulary for unified supervision and enabling autoregressive generation of pixel-level image tokens alongside textual tokens. Our study adopts SigLIP-2~\cite{tschannen2025siglip2multilingualvisionlanguage} as the visual encoder and Qwen-2.5~\cite{qwen2025qwen25technicalreport} as the LLM backbone, trained on 1.04 trillion tokens. UVU significantly mitigates orthogonality between image and text loss gradients, effectively synergizing pixel-level visual perception with semantic-level visual understanding, internalizing visual reconstruction capabilities, markedly enhancing fine-grained visual perception, and unlocking the facilitative role of visual supervision in boosting understanding. Extensive experiments across multiple tasks demonstrate that MLLMs are capable of achieving superior multimodal understanding performance under the supervised learning paradigm of UVU.

In summary, our contributions are as follows:
\begin{itemize}
    \item We demonstrate that end-to-end, unified vision-language supervision during pretraining can fundamentally bolster multimodal understanding. We propose UVU, a novel unified autoregressive framework that bypasses the traditional dependence on VQ-based tokenization. By unifying visual and linguistic generation within a raw, high-dimensional latent space governed by input symmetry, UVU represents the successful realization of visual generative supervision acting as a primary catalyst for enhancing multimodal comprehension.
    \item We analyze the causes of visual supervision impairing multimodal understanding performance in current vision-language unified autoregressive paradigms, attributing it to losses from input visual feature discretization, orthogonality conflicts in image-text loss gradients due to dimensional inconsistencies between visual and language generation, and externalization of visual reconstruction capabilities from reliance on additional visual decoders.
    \item  We introduce a large-scale iterative hierarchical clustering algorithm and construct a 200,000 size pixel-level visual codebook, supporting direct usage, domain-specific fine-tuning, and continuous incremental optimization.
    \item Our extensive experimental results demonstrate that the proposed UVU significantly elevates multimodal understanding capabilities.
\end{itemize}

\section{Related Work}
\label{sec:Related Work}

\subsection{Multimodal Understanding}
To achieve multimodal understanding, MLLMs such as Seed1.5-VL, GLM-4.5, Qwen-VL, and InternVL~\cite{liu2023visualinstructiontuning, bai2023qwenvlversatilevisionlanguagemodel, chen2024internvlscalingvisionfoundation, guo2025seed15vltechnicalreport, 5team2025glm45agenticreasoningcoding} employ pre-trained visual encoders like CLIP or SigLIP~\cite{zhai2023sigmoidlosslanguageimage, radford2021learningtransferablevisualmodels} to extract continuous visual features from images. These encoders leverage contrastive learning on image-text pairs to align visual and textual representations. The visual embeddings are then projected to align with language embeddings and fed into the LLM for text generation. However, this paradigm relies on text-dominant supervision during training, resulting in sparse learning of visual knowledge.

\subsection{Visual Supervision in MLLMs}

Recent research has explored incorporating visual supervision into MLLMs to enhance their visual understanding capabilities. One line of work introduces visual supervision via vision–text alignment~\cite{li2025vistaenhancingvisiontextalignment}, while another focuses on predicting or reconstructing image tokens produced by a vision tokenizer~\cite{wang2024reconstructivevisualinstructiontuning, yoon2025visualrepresentationalignmentmultimodal, li2025spatialforcingimplicitspatial, li2025unleashingintrinsicvisualrepresentation, wang2026autoregressivesemanticvisualreconstruction}. Specifically, Li et al.~\cite{li2025vistaenhancingvisiontextalignment} introduces visual supervision by enforcing alignment between visual and textual representations during training. Wang et al.~\cite{wang2024reconstructivevisualinstructiontuning} introduce visual supervision by employing a Diffusion Transformer (DiT) to predict image features, using the LLM’s output hidden states as the guiding condition. Yoon et al.~\cite{yoon2025visualrepresentationalignmentmultimodal} extract visual features from the intermediate layers of LLM and align them with those from Visual Foundation Models (VFMs) to provide direct visual supervision. Li et al.~\cite{li2025unleashingintrinsicvisualrepresentation} facilitate the learning of more discriminative visual representations via masked image modeling (MIM) within the joint latent semantic space of the LLM.

These paradigms predominantly impose visual supervision during post-training, by which time the backbone's visual representations have already been largely shaped through extensive pre-training. As a result, visual signals function mainly as external constraints or fine-tuning objectives, rather than as intrinsic drivers that can fundamentally reshape fine-grained perceptual representations. This limitation reduces their overall effectiveness. More importantly, it leaves a critical gap: the potential of vision-grounded supervision to guide the formation of visual features during the initial pre-training stage remains largely unexplored.
\begin{figure}[htbp]
  \centering
  \includegraphics[width=1.0\linewidth]{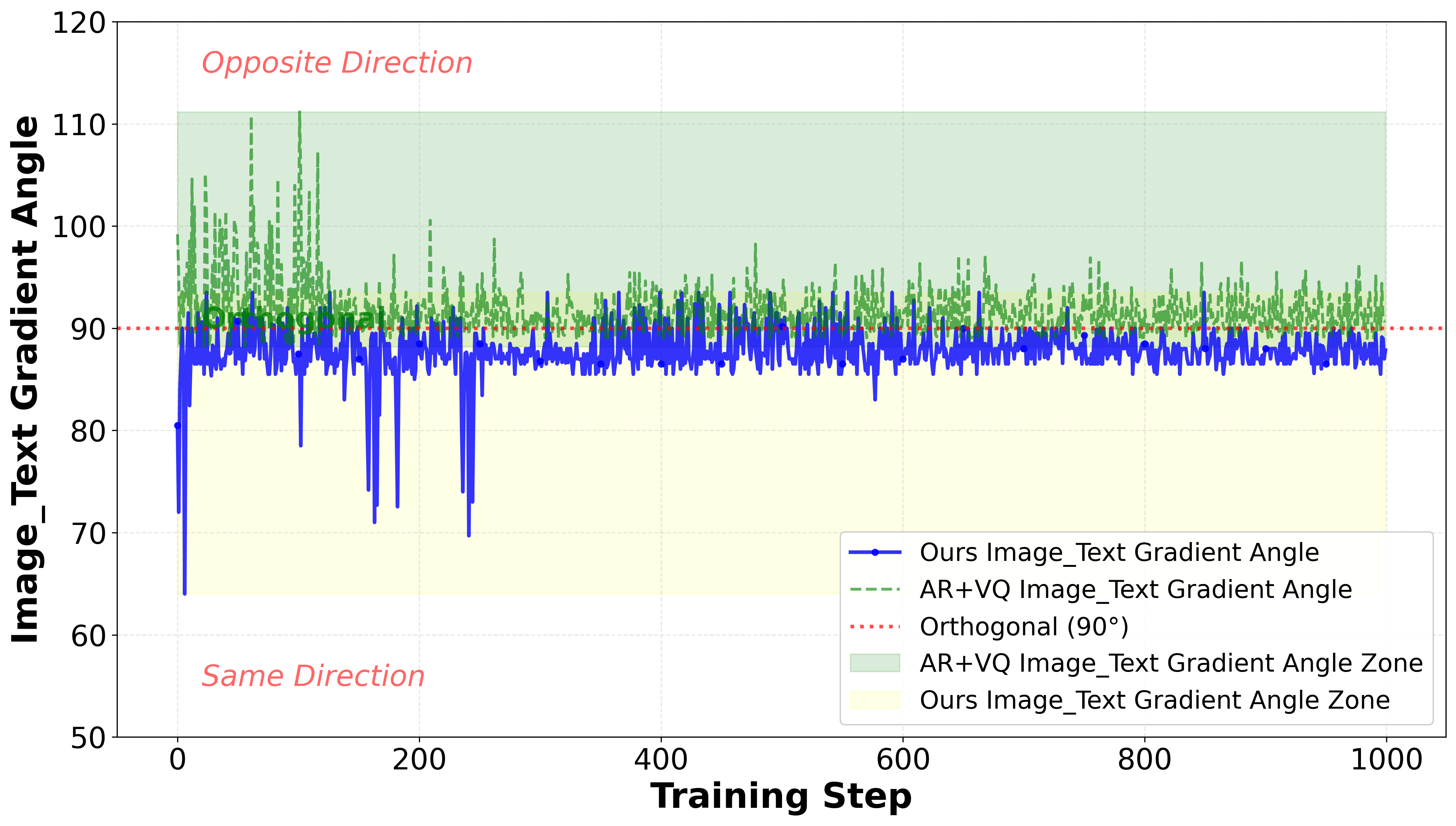}
  \caption{\textbf{The angle between image and text loss gradients over training steps.} The AR+VQ baseline (green) consistently exceeds 90°, indicating orthogonal or opposing directions that hinder multimodal alignment, whereas our UVU (blue) maintains angles below 90°, enabling synergistic optimization.}
  \label{method_1}
\end{figure}

\section{Why Does Visual Supervision Impair Multimodal Understanding?}
To investigate the impact of visual supervision on multimodal understanding within existing AR + VQ frameworks, we construct an autoregressive framework that employs a cascaded VQVAE~\cite{oord2018neuraldiscreterepresentationlearning} as the visual encoder/decoder. As mentioned in Section~\ref{intro}, we observe three primary issues in this framework: 1) discretization of input visual features leading to visual information loss; 2) dimension misalignment dilemmas between low-dimensional visual generation and high-dimensional language generation; and 3) externalization of visual reconstruction capabilities due to reliance on external visual decoders. We consider issues 1) and 3) as deterministic problems induced by structural design and focus on issue 2), analyzing the interaction between visual and linguistic modalities during training through the orthogonality of gradients backpropagated from image and text losses. We quantify this orthogonality as the angle between the image and text loss gradients, formulated as:
\begin{equation}
\cos \theta = \frac{\nabla_i \cdot \nabla_t}{\|\nabla_i\| \|\nabla_t\|},
\end{equation}
where \(\nabla_i\) denotes the gradient from the image loss, \(\nabla_t\) denotes the gradient from the text loss, and \(\theta\) is the angle between them.

As illustrated in the Figure~\ref{method_1}, the angle between the image and text loss gradients consistently exceeds or equals 90 degrees, indicating orthogonal or even opposing directions that result in conflicting optimization paths: the image loss gradients prioritize low-level reconstruction while neglecting global semantics, whereas the text loss gradients promote linguistic coherence but fail to bridge cross-modal representation misalignments. This leads to shifts in visual attention and a degradation in multimodal understanding capabilities. To address these issues, we propose the UVU framework, whose detailed design is elaborated in the subsequent sections.

\section{Method}
In this section, we will introduce our proposed UVU, including the construction of its unique pixel-level visual codebook and the novel Vision-Language Unified Autoregressive framework.

\subsection{Pixel-Level Visual Codebook}

\begin{figure*}[t!]
  \centering
  \includegraphics[width=1.0\linewidth]{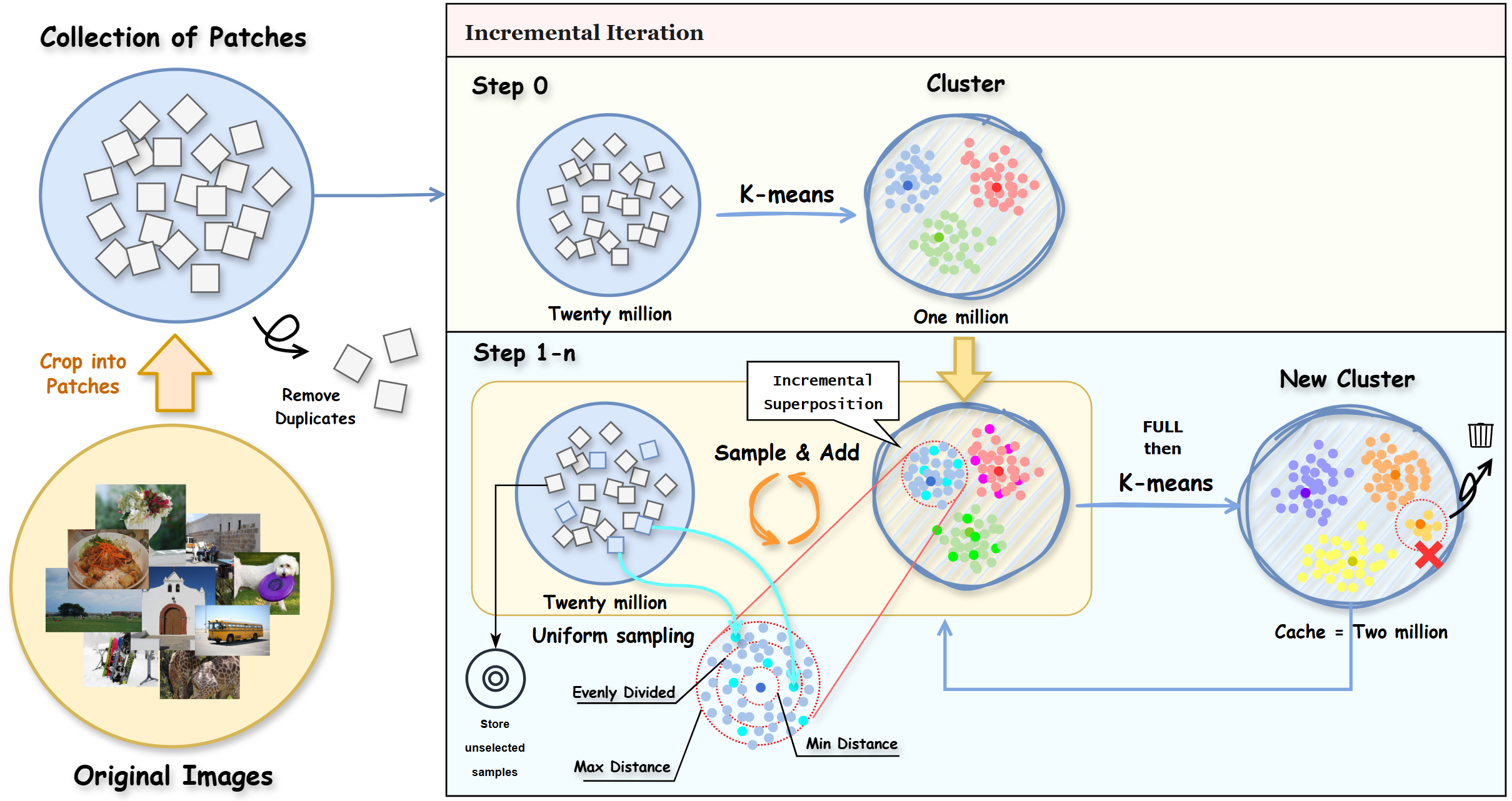}
  \caption{\textbf{Illustration of the large-scale iterative hierarchical clustering algorithm for constructing a pixel-level visual codebook from image patches}. Featuring initial K-means clustering, duplicate removal, uniform sampling, incremental supersampling, and iterative refinement with caching for high coverage and precision.}
  \label{cluster}
\end{figure*}

\begin{figure}[htbp]
  \centering
  \includegraphics[width=1.0\linewidth]{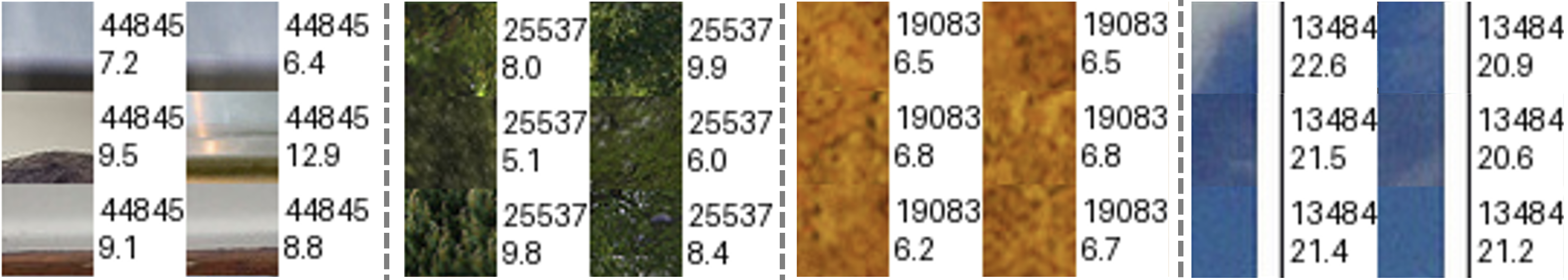}
  \caption{\textbf{Illustration of the clustering visualization for four image token IDs.} For each patch, the values on the right, from top to bottom, represent the image token ID and the distance to the cluster center, respectively.}
  \label{cluster_result}
\end{figure}

To optimize the impact of visual supervision on MLLMs, we eschew reliance on external image generation experts and instead internalize visual reconstruction capabilities within a unified vision-language autoregressive paradigm. Thus, the key challenge lies in constructing an effective visual codebook to extend the vocabulary. Unlike low-dimensional features quantized via VQ, we observe that pixel-level image patches and textual tokens coexist in raw high-dimensional spaces with inherent input symmetry, providing a natural foundation for resolving gradient orthogonality issues arising from vision-language dimensional misalignment. Motivated by this, we propose clustering pixel-level image patches and utilizing the resulting cluster centers to build a pixel-level visual codebook.

To balance inference latency in MLLMs with the fine-grained visual representational capacity of each codeword, we flatten 32$\times$32$\times$3 pixel-level image patches into a 3072-dimensional space as the codeword dimensionality in the pixel-level visual codebook. According to natural image manifold theory~\cite{zhu2018generativevisualmanipulationnatural}, natural images reside on low-dimensional manifolds rather than uniformly populating the high-dimensional pixel space. Specifically, among the $256^{32\times32\times3}$ possible pixel combinations, visually lossless natural image patches occupy only approximately $2^{10}$ potential variants based on manifold dimensionality analysis. Furthermore, considering the limited visual representational capacity of low-resolution images, we focus on precise visual perception for images at resolutions of 512$\times$512 or higher, constructing a pixel-level visual codebook with 200K codewords.

To quantify the quality of the pixel-level visual codebook, we introduce two metrics: Pixel Space Coverage (PSC) and Pixel Representation Precision (PRP), which evaluate and control codebook quality from complementary dimensions. Given a pixel-level visual codebook $\mathcal{C} = \{c_k\}_{k=1}^K$ (where $c_k \in \mathbb{R}^{3072}$ is the $k$-th codeword and $K=2\times 10^5$), and a sample set $\mathcal{P} = \{p_i\}_{i=1}^{N=10^7}$ ($p_i \in \mathbb{R}^{3072}$ represents a reshaped image patch), the Pixel Space Coverage is defined as the minimum of the maximum distances within each cluster divided by the number of active cluster centers (i.e., clusters with at least one assigned sample) in this clustering iteration, serving as a proxy for the codebook's efficient span over the natural image manifold:
\begin{equation}
\text{PSC} = \frac{\min_{k=1,\dots,K} \left( \max_{p_i \in \mathcal{S}_k} \|p_i - c_k\|_2 \right)}{K_{\text{active}}},
\end{equation}
where $\mathcal{S}_k = \{p_i \mid k = \arg\min_j \|p_i - c_j\|_2\}$ denotes the set of samples assigned to the $k$-th cluster, and $K_{\text{active}} = |\{k \mid |\mathcal{S}_k| > 0\}|$ is the count of active clusters. The Pixel Representation Precision quantifies reconstruction accuracy on the sample set $\mathcal{P} = \{p_i\}_{i=1}^{N=10^7}$, where for each image patch $p_i$, we assign it to the nearest codeword $\hat{p}_i = c_{k_i}$ (with $k_i = \arg\min_k \|p_i - c_k\|_2$) and compute the complement of the normalized mean squared error (MSE):
\begin{equation}
\text{PRP} = 1 - \frac{1}{M} \sum_{m=1}^M \frac{\| \hat{p}_m - p_m \|_2^2}{\sigma_{\max}^2},
\end{equation}
where $\sigma_{\max}^2$ is the global maximum variance across the test set for normalization. Our objective is to maximize these metrics, ensuring broad coverage of the natural image space while achieving precise intra-cluster representations. The joint optimization target is formulated as:
\begin{equation}
\max_{\mathcal{C}} \left( \text{PSC} + \text{PRP} \right).
\end{equation}

To achieve this objective, we propose a large-scale iterative hierarchical clustering algorithm that optimizes coverage and precision through image patch deduplication, hierarchical sampling, and iterative filtering. The process is illustrated in the Figure~\ref{cluster}. We begin by cropping 2 billion 32$\times$32$\times$3 image patches from 1 million images and reshaping them into 3072-dimensional feature vectors $\{p_i\}_{i=1}^{2\times10^9}$. Given that high-dimensional feature clustering relies solely on L2 distance metrics, it often struggles to distinguish patches with similar color distributions but distinct spatial configurations, leading to the loss of structural continuity. Furthermore, raw pixel vectors lack an explicit spatial inductive bias, causing the resulting codebook to capture isolated textures rather than coherent perceptual units. To resolve this, we first apply sinusoidal positional encoding to the 3072-dimensional features to incorporate spatial positional information:
\begin{equation}
p_i[j] = p_i[j] +
\begin{cases}
\sin\left(\frac{j}{10000^{2j/d}} \right) & \text{for even } j, \\
\cos\left(\frac{j}{10000^{2(j-1)/d}} \right) & \text{for odd } j.
\end{cases}
\end{equation}

where $d=3072$ is the dimensionality, and $j$ indexes the dimension. To mitigate catastrophic shifts in clustering caused by duplicate image patches, we randomly initialize 200K cluster centers $\{c_k^{(0)}\}$ and assign each feature to its nearest center:

\begin{equation}
k_i = \arg\min_k \|p_i - c_k^{(0)}\|_2.
\end{equation}

Within each cluster, we compute distances $d_{i,k} = \|p_i - c_k\|_2$ (precise to five decimal places) and discard features with duplicate distances, retaining only unique samples for the initial K-means clustering:

\begin{equation}
\min_{\{c_k\}} \sum_{i=1}^N \|p_i - c_{k_i}\|_2^2.
\end{equation}

Upon obtaining the initial cluster centers, we incorporate additional data into an iterative process: in each iteration, we sample 20 million image patches and assign them to the corresponding centers. We first compute the sample count $|\mathcal{S}_k|$ for each cluster, filter out long-tail clusters where $|\mathcal{S}_k| < \delta$ (and cache their centers and features), and for the remaining clusters, compute distances, discard duplicates, and maintain a global distance dictionary $\mathcal{D} = \{d_{k,j}\}$ ($j$ indexes unique distances within the cluster). We then perform hierarchical sampling on the effective distance distributions to ensure uniform distribution of training samples. The hierarchically sampled and long-tail filtered samples are cached until the accumulated sample size reaches $\alpha$, at which point we re-cluster using the cached features to update the centers and refresh the distance dictionary. This process iterates until convergence, formulated as:

\begin{equation}
c_k^{(t+1)} = \frac{1}{|\mathcal{S}_k^{(t)}|} \sum_{p_i \in \mathcal{S}_k^{(t)}} p_i,
\end{equation}
where $t$ denotes the iteration step, the visualization of the clustering results is illustrated in the Figure~\ref{cluster_result}.

This large-scale iterative hierarchical clustering algorithm is implemented based on Faiss~\cite{johnson2017billionscalesimilaritysearchgpus}, leveraging distributed streaming computation for efficiency. Large-scale indexing slices the data and distributes it across sub-ranks for execution, synchronizing back to the main rank for unified training, achieving high efficiency, low memory consumption, and scalability. Notably, this algorithm not only continuously optimizes the pixel space coverage and representation precision of the pixel-level visual codebook but also provides an incremental interface for ongoing refinement with new data samples. Moreover, it can be extended to domain-specific areas to deepen fine-grained representations.

\subsection{Vision-Language Unified Autoregressive}
\begin{figure*}[t!]
  \centering
  \includegraphics[width=0.89\linewidth]{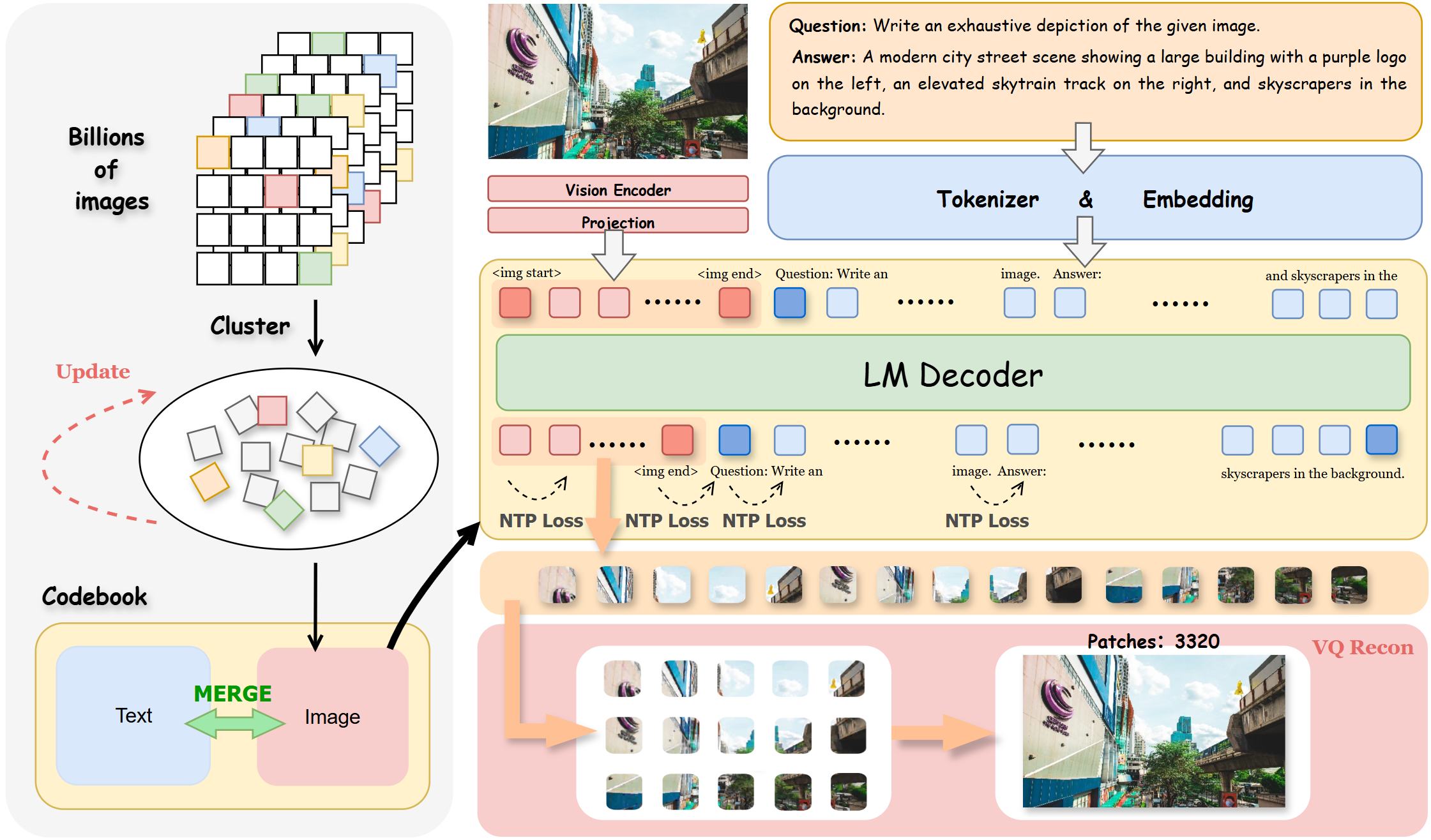}
  \caption{\textbf{Overview of the UVU vision-language unified autoregressive framework.} Continuous visual features from SigLIP-2 are projected and integrated with textual embeddings for autoregressive next-token prediction in an LM decoder, generating pixel-level image tokens indexed via a clustered codebook for lossless reconstruction, merging text and image modalities.}
  \label{pipeline}
\end{figure*}

Building upon the pixel-level visual codebook constructed in the previous section, we extend the vocabulary to establish the UVU framework, a unified autoregressive paradigm for vision and language. As illustrated in the Figure~\ref{pipeline}, textual token sequences are transformed into textual embedding sequences via an embedding layer. Leveraging the inherent input symmetry between pixel-level image patches and textual tokens in raw high-dimensional spaces, we eliminate the need for discretization of input visual features and external visual decoders at the output. In the visual modality, continuous visual features are extracted using SigLIP-2, projected through a projection layer, and seamlessly integrated with the textual embedding sequences without loss. These combined embeddings are then fed into a large language model (LLM) decoder for autoregressive next-token prediction, enabling the generation of both image tokens and text tokens. This process is formalized as:
\begin{equation}
\mathcal{L} = 0.5*\mathcal{L}_{\text{image}} + \mathcal{L}_{\text{text}},
\end{equation}
where $\mathcal{L}_{\text{image}} = -\sum_{i} \log P(t_i^{\text{image}} \mid t_{<i})$ and $\mathcal{L}_{\text{text}} = -\sum_{i} \log P(t_i^{\text{text}} \mid t_{<i})$, with \(t_i^{\text{image}}\) and \(t_i^{\text{text}}\) representing the target image and text tokens at position \(i\), conditioned on preceding tokens \(t_{<i}\).

The output visual tokens, aligned with textual tokens, are indexed through the pixel-level visual codebook to corresponding pixel-level image patches. We observe that image tokens learned via this autoregressive approach inherently incorporate spatial information within the generated image patches and support dynamic resolution generation. Analyzing the frequency distribution of tokens corresponding to 500 million image patches in the training data reveals a pronounced long-tail distribution. Through visualization, high-frequency image token IDs are found to typically correspond to low-degree-of-freedom background regions, introducing redundancy into the training process. To promote training diversity, we compute the reciprocal of frequencies from 5 billion image token ID samples as sampling probabilities, thereby equalizing learning probabilities across image token IDs and ensuring uniform learning of image patches. This strategy is formalized as:

\begin{equation}
p_k = \frac{1 / f_k}{\sum_{j=1}^K 1 / f_j},
\end{equation}
where \(f_k\) is the frequency of the \(k\)-th image token ID, and \(K\) is the codebook size.

During the pre-training phase, to maintain balance between image and text tokens in supervised training, we apply secondary random dynamic sampling to the image token IDs in each batch, ensuring equilibrated multimodal learning:

\begin{equation}
\text{Image token ID}_b \sim \text{Uniform}(\{k \mid p_k > \tau_b\}),
\end{equation}
where \(\tau_b\) is a batch-specific threshold dynamically adjusted to balance token counts. In the subsequent supervised fine-tuning (SFT) phase, we discontinue loss supervision on image tokens to achieve more precise instruction comprehension.

\section{Experiment}

\subsection{Experimental Setup}

\textbf{Training Details.}
(1) \textbf{Training Data:} Throughout the entire training pipeline, encompassing both pre-training and post-training stages, we utilize a total of 1.04 trillion tokens sourced from open datasets including LLaVA-OneVision~\cite{li2024llavaonevisioneasyvisualtask}, FineVision~\cite{wiedmann2025finevisionopendataneed}, Cauldron~\cite{laurençon2024mattersbuildingvisionlanguagemodels}, and Cambrian-7M~\cite{tong2024cambrian1fullyopenvisioncentric}, as well as our proprietary dataset. These datasets encompass three modalities: Text-Only, Text-Image, and Text/Image Interleaved.
(2) \textbf{Model Components:} We employ Qwen2.5-3B-Instruct as the language backbone and SigLIP2-so400m-patch16-naflex as the vision encoder.
(3) \textbf{Codebook Construction:} The codebook, comprising 200K entries, is derived from 2 billion 32$\times$32$\times$3 patches using our proposed hierarchical iterative clustering algorithm.
(4) \textbf{Training Setup:} Training is conducted with a per-GPU batch size of 1 with sequence length 32K. We adopt the AdamW optimizer with a learning rate of 2e-5 and a warmup ratio of 0.01.

\textbf{Evaluation Details.}
We evaluate the multimodal understanding performance of UVU on a diverse set of vision-centric vision-language benchmarks, including SEED-Bench~\cite{li2023seedbenchbenchmarkingmultimodalllms}, VisuLogic~\cite{xu2025visulogicbenchmarkevaluatingvisual}, MMStar~\cite{chen2024rightwayevaluatinglarge}, MME~\cite{fu2025mmecomprehensiveevaluationbenchmark}, CVBench-2D, CVBench-3D~\cite{tong2024cambrian1fullyopenvisioncentric}, VLMBlind~\cite{rahmanzadehgervi2025visionlanguagemodelsblind}, RefCOCO~\cite{yu2016modelingcontextreferringexpressions}, LISA-Grounding~\cite{lai2024lisareasoningsegmentationlarge}, ScienceQA~\cite{lu2022learnexplainmultimodalreasoning}, BLINK~\cite{fu2024blinkmultimodallargelanguage}, and HallusionBench~\cite{guan2024hallusionbenchadvanceddiagnosticsuite}.

\subsection{Ablation Study}

\textbf{Ablation Study on visual supervision}. As shown in Table~\ref{ablation}, we compare the performance of models trained under identical training data and experimental settings using without visual supervision, AR+VQ visual supervision, and our UVU approach. The results indicate that the AR+VQ scheme substantially degrades the model's visual understanding capabilities. In contrast, UVU enhances visual understanding, with particularly pronounced improvements on tasks requiring strong visual perception, such as the visual grounding benchmark RefCOCO. This demonstrates the benefits that elevated visual perceptual abilities bring to overall visual comprehension.

\setlength{\tabcolsep}{3.5pt}
\begin{table}[htbp]
\centering
\small
\adjustbox{max width=1.0\linewidth}{
\begin{tabular}{l | c c c }
\toprule
\textbf{Method} & \textbf{MMStar} & \textbf{RefCOCO} & \textbf{MMB} \\
\midrule
wo/visual supervision & 52.9 & 85.6 & 74.3 \\
AR+VQ & 46.5 & 79.6 & 68.2 \\
\rowcolor{gray!20} \textbf{Ours} & 55 & 91.8 & 76.1 \\
\bottomrule
\end{tabular}
}
\caption{Ablation study on different visual supervision.}
\label{ablation}
\end{table}

\textbf{Ablation study on visual codebook size.} To validate the impact of pixel-level visual codebook size, we evaluate PSC and PRP metrics, along with performance on visual understanding benchmarks, under identical training data and experimental settings across varying codebook sizes. As shown in the Table~\ref{codebook}, both PRP and understanding performance steadily improve with increasing codebook size, demonstrating that larger codebooks enhance fine-grained visual representation and reconstruction fidelity, which in turn drives gains in downstream test performance. This confirms that PixelUnd effectively leverages visual supervision to boost understanding and that finer-grained supervision enables the model to acquire more precise visual perception, thereby elevating multimodal understanding capabilities. Additionally, PSC remains stable across sizes, underscoring the robustness of our large-scale iterative hierarchical clustering algorithm. Given that 500K offers no substantial gains over 200K, we select 200K as the final codebook size to balance reconstruction quality and inference efficiency.

\setlength{\tabcolsep}{3.5pt}
\begin{table}[htbp]
\centering
\small
\adjustbox{max width=1.0\linewidth}{
\begin{tabular}{l | c c c c}
\toprule
\textbf{codebook size} & \textbf{MMB} & \textbf{RefCOCO} & \textbf{PSC} & \textbf{PRP (\%)} \\
\midrule
50k  & 74.7 & 88.7 & 0.336 & 86.42 \\
100k & 75.2 & 90.1 & 0.344 & 91.26 \\
\rowcolor{gray!20} \textbf{200k} & 76.1 & 91.8 & 0.341 & 95.63 \\
500k & 76.4 & 91.7 & 0.337 & 96.04 \\
\bottomrule
\end{tabular}
}
\caption{Ablation study on visual codebook size.}
\label{codebook}
\end{table}

\textbf{Ablation on large-scale iterative hierarchical clustering algorithm}
As shown in the Table~\ref{ihc}, we conduct ablation studies on two critical components of our pixel-level visual codebook construction: positional encoding in the feature dimension and the iterative hierarchical clustering algorithm. The results clearly demonstrate that each component contributes significantly to the overall performance, validating their essential roles in achieving high-quality codebook representation and downstream multimodal understanding gains.

\setlength{\tabcolsep}{3.5pt}
\begin{table}[htbp]
\centering
\small
\adjustbox{max width=1.0\linewidth}{
\begin{tabular}{l | c c c c}
\toprule
\textbf{Method} & \textbf{PSC} & \textbf{PRP (\%)} & \textbf{RefCOCO} & \textbf{MMB} \\
\midrule
K-means & 0.014 & 70.14 & 81.0 & 66.3 \\
+ position encode & 0.133 & 82.84 & 86.9 & 71.1 \\
\rowcolor{gray!20}\textbf{+ iterative hierarchical clustering} & 0.341 & 95.63 & 91.8 & 76.1 \\
\bottomrule
\end{tabular}
}
\caption{Ablation with different codebook construction methods.}
\label{ihc}
\end{table}

\subsection{Qualitative Results}

\begin{figure*}[htbp]
  \centering
  \includegraphics[width=1.0\linewidth]{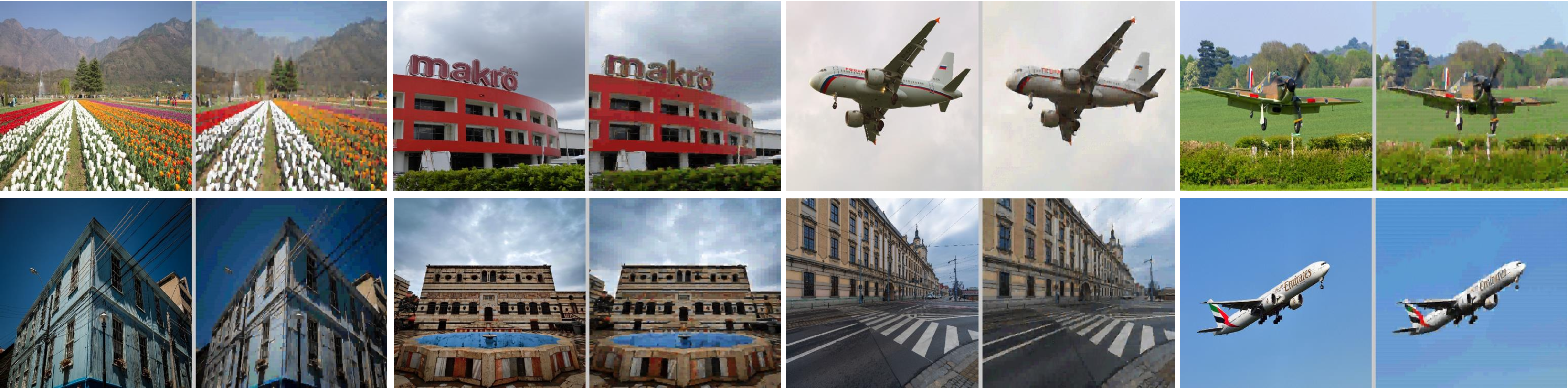}
  \caption{\textbf{Original images and their reconstructions by UVU.} Each pair consists of the original image on the left and the reconstructed version on the right.}
  \label{reconstruct}
\end{figure*}

\begin{figure}[htbp]
  \centering
  \includegraphics[width=1.0\linewidth]{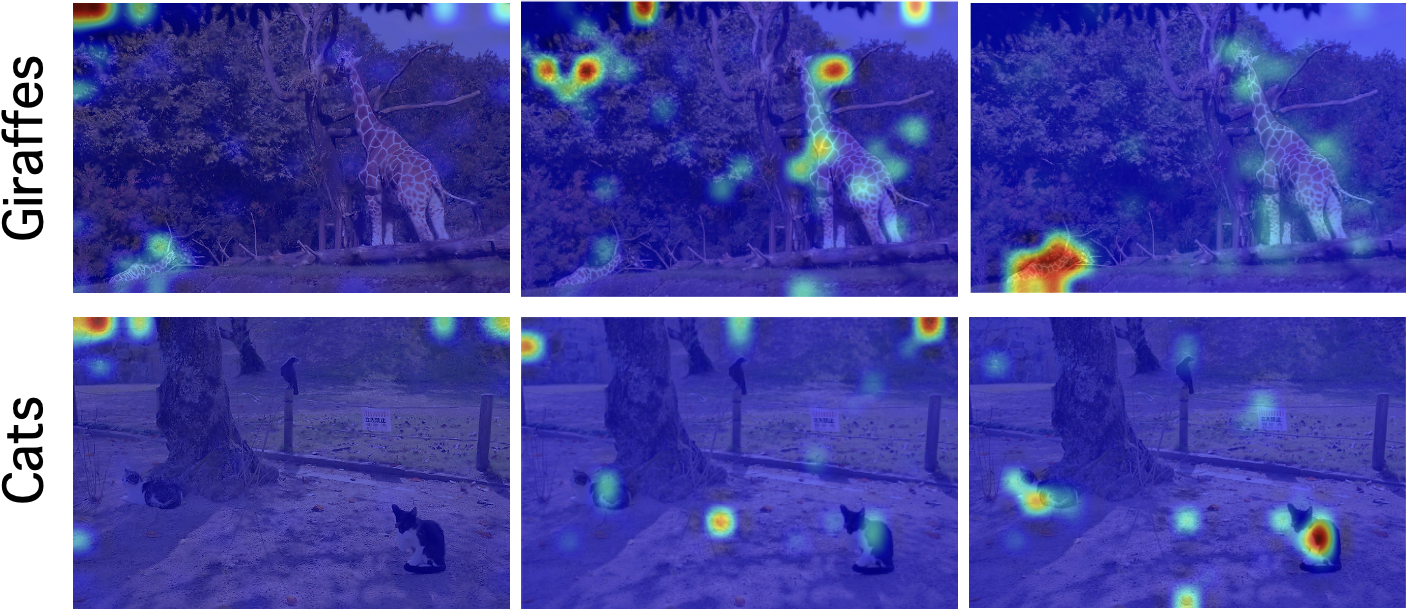}
  \caption{\textbf{Comparison of visual attention heatmaps under three paradigms}: from left to right, without visual supervision, AR + VQ, and UVU (Ours). The query is "How many [objects] are in the image?" with the specific objects (giraffes or dogs) indicated on the left of each example. }
  \label{attention}
\end{figure}

\setlength{\tabcolsep}{3.5pt}
\begin{table*}[htbp]
\centering
\small
\adjustbox{max width=\textwidth}{
\begin{tabular}{l | c | c | c c c c c c c c c c c c }
\toprule
\textbf{Method} & \textbf{\# Params} & \textbf{LLM Backbone} & \textbf{SEEDB} & \textbf{VisuLogic} & \textbf{MMStar} & \textbf{MME} & \textbf{CVB2D} & \textbf{CVB3D} & \textbf{VLMBlind} & \textbf{RefCOCO} & \textbf{LISA} & \textbf{ScienceQA} & \textbf{BLINK} & \textbf{HallusionB} \\
\midrule
\multicolumn{15}{c}{\textcolor[RGB]{105, 105, 105}{\textit{Wo/Visual Supervision}}} \\
\midrule
Qwen2.5-VL & 3B & Qwen2.5 & 73.3 & 22.8 & 52.8 & 2141.7 & 71.1 & 71.9 & 34.6 & 84.1 & 57.4 & 79.2 & 46.9 & 64.5 \\
GLM-4v & 9B & GLM-4 & 71.1 & -- & 54.8 & 2018.8 & 63.2 & 63.4 & -- & -- & -- & \textbf{96.7} & -- & 45 \\
InternVL2 & 2B & InternLM2 & 70.9 & 18.7 & 49.8 & 1864.3 & 61.4 & 61.3 & 29.7 & 77.8 & 46.3 & 94.1 & 42.8 & 38 \\
LLaVA-OV & 7B & Qwen2 & \textbf{75.4} & 25.3 & \textbf{56.7} & 2146.3 & 62.3 & 63.4 & -- & 78.1 & 47.4 & 86.6 & 46.1 & 47.5 \\
DeepSeek-VL & 7B & DeepSeek-LLM & 70.1 & -- & 40.5 & 1765.4 & -- & -- & -- & -- & -- & 80.9 & 40.9 & 34.5 \\
ShareGPT4V & 13B & Vicuna-13B & 70.6 & -- & 38.3 & 1914.9 & -- & -- & -- & -- & -- & 69.5 & 40.9 & 28.4 \\
LLaVA-v1.5 & 13B & Vicuna-13B & 68.2 & 24.6 & 34.3 & 1780.8 & 53.1 & 53.3 & -- & 73.5 & 40.4 & 72.6 & 40.9 & 24.5 \\
\midrule
UVU* & 3B & Qwen2.5 & 73.1 & 24.3 & 52.9 & 2056.6 & 68.4 & 67.8 & 33.4 & 85.6 & 59.6 & 88.4 & 46.1 & 59.2 \\
\rowcolor{gray!20} \textbf{UVU(ours)} & 3B & Qwen2.5 & 74.1 & \textbf{26.4} & 55.0 & \textbf{2201.9} & \textbf{73.0} & \textbf{76.6} & \textbf{38.8} & \textbf{91.8} & \textbf{71.7} & 91.3 & \textbf{52.8} & \textbf{66.6} \\
\bottomrule
\end{tabular}
}
\caption{Evaluation on multimodal understanding benchmarks. UVU* refers to the version without visual supervision.}
\label{table_und}
\end{table*}

\textbf{Reconstruction Results.}
As illustrated in Figure~\ref{reconstruct}, we present comparisons between original images and their reconstructions generated during the training process. Notably, UVU achieves reconstruction of visual details without relying on any external visual decoders. This outcome further validates the model's ability to learn fine-grained visual knowledge, including perceptual and spatial understanding, thereby significantly enhancing its overall visual understanding performance.

\textbf{Visual Attention Map Analysis.}
As depicted in Figure~\ref{attention}, we compare the visual attention heatmaps produced during model inference. It is evident that UVU's attention is more precisely concentrated on semantically relevant regions compared to baselines, underscoring its superior fine-grained visual perception capabilities.

\subsection{Quantitative Results}

As shown in Table~\ref{table_und}, we evaluate the multimodal understanding capabilities of UVU on 12 authoritative multimodal understanding datasets. Our observations are as follows: (1) Relative to leading models trained solely with textual supervision, UVU demonstrates substantial improvements, as exemplified by its performance over Qwen2.5-VL; (2) Benefiting from UVU's robust capture and learning of fine-grained visual knowledge, \textbf{it exhibits particularly outstanding results on tasks that emphasize visual perception}, such as RefCOCO, LISA, CVBench, and BLINK, thereby validating its strong visual perceptual abilities. Furthermore, given UVU's inherent image generation capabilities, we anticipate even greater enhancements in future iterations through the incorporation of image generation data during training.

\section{Conclusion}
In this work, we observed that efforts to integrate visual supervision in unified autoregressive multimodal models, aimed at bridging understanding and generation, have compromised multimodal understanding performance due to challenges like visual feature discretization and gradient orthogonality between image and text losses. By recognizing the shared raw high-dimensional space and input symmetry between pixel-level image patches and textual tokens, we propose \textbf{UVU},  innovatively bypasses vector quantization, utilizing continuous visual encoding for lossless inputs and a scalable iterative hierarchical clustering algorithm to forge a pixel-level visual codebook. This enables seamless autoregressive generation of pixel-level image and textual tokens, harmonizing pixel-level perception with semantic understanding, internalizing generation capabilities, and achieving enhancement of comprehension through visual supervision. Rigorous evaluations across diverse tasks affirm that UVU surpasses SOTA MLLMs in multimodal understanding.

\clearpage

\section*{Acknowledgements}
This work was partly supported by the Special Foundations for the Development of Strategic Emerging Industries of Shenzhen(No.KJZD20231023094700001).

{
    \small
    \bibliographystyle{ieeenat_fullname}
    \bibliography{main}
}

% \clearpage


% \clearpage

% {
%     \small
%     \bibliographystyle{ieeenat_fullname}
%     \bibliography{main}
% }


\end{document}